\documentclass[11pt]{article}

\usepackage[T1]{fontenc}
\usepackage[utf8]{inputenc}
\usepackage[margin=1in]{geometry}
\usepackage{amsmath,amssymb,amsfonts,amsthm}
\usepackage{mathrsfs}
\usepackage{enumitem}
\usepackage[protrusion=true,expansion=false]{microtype}
\usepackage[hidelinks]{hyperref}

\theoremstyle{plain}
\newtheorem{theorem}{Theorem}

\newtheorem{corollary}[theorem]{Corollary}
\newtheorem{proposition}[theorem]{Proposition}

\theoremstyle{definition}
\newtheorem{definition}{Definition}
\newtheorem{thesis}{Thesis}
\newtheorem{principle}{Principle}

\theoremstyle{remark}
\newtheorem{remark}[theorem]{Remark}
\newtheorem{researchproblem}{Open Problem}

\newcommand{\Heven}{\mathcal{H}_{\mathrm{even}}}
\newcommand{\Hodd}{\mathcal{H}_{\mathrm{odd}}}
\newcommand{\im}{\operatorname{im}}
\newcommand{\R}{\mathbb{R}}
\newcommand{\C}{C_\bullet}
\newcommand{\dpar}{\partial}
\newcommand{\kap}{\kappa}

\title{\textbf{Computation, Condensation, and the Incompleteness\\
Between Them: A Coupled Foundation of Intelligence}}

\author{Xin Li\\
Department of Computer Science, University at Albany, SUNY\\
\texttt{xli48@albany.edu}}

\date{Position paper, July 2026}

\begin{document}
\maketitle

\begin{abstract}
The theory of computation was built to answer Turing's question: what is
effectively calculable by an unbounded, immortal, disembodied agent following
rules? Intelligence answers a different question (nature's): what can a
\emph{finite}, mortal, energy-limited agent do quickly enough to survive in a
non-stationary world? We argue that a complete answer requires two
operators: \emph{computation} and \emph{memorizaion}. Computation, $\dpar$, transforms structure toward closure;
memorization, $\kap$, condenses a validated closed cycle into a reusable token.
Turing formalized $\dpar$ and abstracted $\kap$ away, because his agent had
infinite time and never needed to amortize. The new insight of our position paper is that the coupling of the two is not optional but \emph{forced}, and forced by a
precise mathematical fact: \textbf{neither operator alone can be complete}. We prove that
symbolic computation confined to a discrete sector suffers G\"odel's
diagonalization incompleteness, that geometric descent confined to a continuous
sector suffers a Morse forced-saddle incompleteness, and that these two are not
analogies but the parity-conjugate faces of a single obstruction on a coherent
complex with $\dpar^2=0$ - the even face realized by diagonalization, the odd
by the topologically forced saddle, and no resolution confined to one parity
able to come full circle. Intelligence must therefore couple both modes. We
then locate the price of the coupling: its hinge operation,
context-identification (the recognize-versus-discover decision), is exactly
where the two incompletenesses coincide, hence undecidable and carrying an
irreducible error floor. Finally we argue that the coupling is a universal law, realized, in the emergent sense of Anderson's ``more is different,'' at
every scale from genes to thoughts to cultures, and give its falsifiable
core and honest scope.
\end{abstract}

%\tableofcontents

% =====================================================================
\section{The Problem: Intelligence Under Finitude}
\label{sec:problem}
%\addcontentsline{toc}{section}{\textbf{Part I. The Problem: Intelligence Under Finitude}}

\subsection{Two questions}
\label{sec:two-questions}

Alan Turing did not set out to define intelligence \cite{turing1950computing}. He set out to define the
\emph{effectively calculable} \cite{turing1936computable}: what an agent with tape, unlimited time, and no
insight could compute by following rules mechanically. The machine he
abstracted from that agent (unbounded tape, no clock, no metabolism, and no
stake in whether it halts) is a monument precisely because it factors
\emph{out} everything contingent about who is doing the computing. The
Church-Turing thesis \cite{sipser_introduction_1996} is the claim that what remains, after finitude and
embodiment and mortality are removed, is a single robust notion of computation.
%This was the right abstraction for its question. It is the wrong abstraction for ours.
This abstraction is exactly what makes Turing computation universal, but it is
also exactly what prevents it from being a theory of biological intelligence \cite{buzsaki2006rhythms}.

Nature never asked what is calculable in principle because it does not have foresight \cite{dawkins1996blind}. Evolution is not a
proof-search over the space of computable functions but a bounded process
selecting, under relentless resource pressure, systems that act well enough,
soon enough, to persist. The question nature poses to any organism is not
``what can be computed?'' but ``what can \emph{you} compute, with the tissue and
the seconds and the calories you actually have, against a world that changes
faster than you can exhaustively model it?'' These are not the same question at
different scales of difficulty. They are different questions, and the second one
has the additional variables (e.g., the agent's time, memory, energy, and the
non-stationarity of its external environment) that the first was expressly built to erase.

The thesis of this paper is that intelligence cannot be characterized by
Turing computation alone \cite{sipser_introduction_1996}, nor by neural condensation alone \cite{hertz_introduction_2018}, but by their
coupling. The Turing machine formalizes one organizational principle:
computation as explicit, rule-governed transformation of states. Neural
systems expose a contrasting principle: condensation, by which finite agents
compress recurrent experience into reusable structure under constraints of
time, memory, energy, and changing environments. Computation answers what can
be derived by procedure; condensation answers what can be retained, recognized,
and amortized by a bounded organism \cite{savitch1970}. A theory of intelligence must move from the isolated problem to the agent embedded in a changing world. Its
question is not simply ``is this computable in principle?'' but ``what can this
system compute and condense quickly enough to matter?'' This relocation does
not weaken the theory but identifies the missing operator. Intelligence, we
will argue, requires the interaction of the Turing operator of computation and
the neural operator of condensation, because the procedure without condensation
cannot amortize experience, while condensation without procedure cannot
generate open-ended computation.

\subsection{Three finitudes, three forced structures}
\label{sec:finitudes}

If intelligence is what finitude forces on computation, the architecture of
intelligence should be \emph{derivable} from the constraints of being a finite
agent, rather than designed \cite{ashby1952design}. We argue it is: three finitudes each force a
structural feature that the resource-unbounded Turing agent has no reason to
possess.

\paragraph{Finite time forces amortization.}
An immortal computer can afford to recompute. If there is always more time, then
having solved a problem confers no advantage the next time it is posed; one
simply solves it again. This is the regime of Savitch's
theorem~\cite{savitch1970}, where reachability is decided in minimal space by
recomputing sub-paths without ever storing them, at the cost of superpolynomial
time (time being free). The Turing machine is, in this exact sense, an
immortal agent: its tape is scratch space, not memory, because nothing it
computes changes what future computation is cheap. A \emph{mortal} agent cannot
live this way: it will die before it re-derives what it already knew; therefore the solved computation is forced to convert into cheap future retrieval (i.e., to
\emph{amortize} \cite{gershman2014amortized}). Amortization is not an optimization a finite agent may adopt but a debt mortality collects. The operator that performs it, storing a solved
structure so the next inference is a lookup rather than a re-derivation, is
absent from the theory of computation for the simple reason that Turing's agent
never dies.

\paragraph{Finite memory forces abstraction.}
An agent with unbounded tape could, in principle, memoize \emph{everything}:
store every configuration it has ever resolved. A finite agent facing an
unbounded stream of experience cannot. The number of distinct situations grows
without bound while its storage does not. It is forced to store at the
\emph{coarsest granularity that preserves what matters}: not configurations
themselves, but \emph{equivalence classes of configurations}, one token per
recurring context. This is abstraction, which is not a cognitive luxury but a
consequence of bounded capacity meeting unbounded experience
\cite{simon1962architecture}. The recursive form, in which tokens are grouped
into higher tokens and contexts into super-contexts, is the only way a
fixed-width memory can absorb complexity that grows without bound: complexity
is serialized into \emph{depth} because it cannot be accommodated in
\emph{width}. The operator that performs this compression, replacing many
resolved particulars by one reusable scaffold, is absent from classical
computation for the same reason amortization is absent: Turing's agent never
has to consolidate. Its tape need not forget, summarize, or sleep on what it
has learned. A biological agent does because sleep is the temporal form of abstraction \cite{hinton1995wake}.

\paragraph{Finite energy and real-time action force the parity split.}
An agent that could stop the world to learn, then resume to act, could keep
learning and acting in separate phases indefinitely. An embodied agent cannot:
it must act \emph{while} it learns, on the same physical substrate, in real
time, and the learning must not destroy the acting. This is the
stability-plasticity problem \cite{mermillod2013stability}, and its solution under a single finite substrate
is a decomposition into two subspaces that do not interfere: one plastic,
where new structure is explored, and one stable, where consolidated structure is
preserved and used. The two must be, in the appropriate sense, orthogonal, so
that change in one does not corrupt the other. A finite agent that must be
simultaneously plastic and stable is forced to partition its state into
complementary sectors and, as we show next, that partition is not just
convenient but topologically forced (it is where incompleteness lives \cite{godel1931,morse1934}).

\medskip
\noindent
None of these three pressures (amortization, abstraction, sector-partition) acts on Turing's agent, because that agent is immortal, unbounded, and
disembodied. This is why none of the corresponding structure appears in the
classical theory of computation \cite{sipser_introduction_1996}. It is not that Turing missed them; he
\emph{abstracted them away}, correctly, for his question. They return, forced
and non-optional, the moment the question becomes nature's. The position we advocate through this paper is to carefully restore the
conditions that Turing machines deliberately removed and ask what new operations become necessary
under them. Once computation is placed back inside a finite, embodied,
non-stationary agent, the missing pressures do not appear as vague biological
complications; they organize themselves into two complementary operations, as we will elaborate next.
%One extends structure by rule-governed transformation. The other compresses resolved structure into reusable form. We call these the two operators.

\subsection{The two operators}
\label{sec:operators}

The three finitudes force a second \emph{operation}, categorically
distinct from computation, that a pure computing engine cannot perform on
itself. To compute is to \emph{elaborate} (i.e., to
build a derivation, extending an unresolved state toward a closed,
self-consistent form); to memorize is to \emph{collapse} (i.e., to take a completed,
closed form and quotient it to a single reusable token, discarding the internal
structure that produced it) \cite{squire_memory_2015}. Elaboration accumulates structure; condensation
strips it away, keeping only the handle. A single operation cannot do both,
because these are inverse motions: one builds up the very structure the other
tears down. %This is the origin of the two operators, and the reason a theory of intelligence must carry a pair rather than a single, more powerful engine.
Both operations live on a common substrate, which we call a \emph{coherent
complex}: a graded space of representable states with a boundary operator
$\dpar$ satisfying the coherence law $\dpar^2=0$ \cite{wheeler1990information}, equipped with a metric. The
coherence law is the constitutive assumption, the condition under which ``a
self-consistent structure'' is well defined at all, in the way the triangle
inequality is the condition under which ``distance'' is well defined. Part~II
makes this complex precise (Definition~\ref{def:complex}); here we name the two
operators it carries.

\paragraph{$\dpar$: computation as transformation toward closure.}
The boundary operator transforms a structure by one step. A computation is a
sequence of such steps that drives an \emph{open} structure ($\dpar z \neq 0$,
an unfinished trajectory, a conjecture with loose ends) toward a \emph{closed}
one ($\dpar z = 0$, a cycle, a self-consistent result). This is exactly what a
Turing machine does \cite{turing1936computable}: it rewrites a configuration, step by step, and
``computation closes'' (it halts) when there is nothing left to transform.
The Turing machine is the $\dpar$-engine: it computes \emph{within a single
parity sector}, driving open chains to closed ones by repeated rule-application.
Its criterion of success is \emph{halting on a correct symbol}.

\paragraph{$\kap$: memorization as condensation across sectors.}
The condensation operator does something $\dpar$ cannot: it takes a
\emph{closed} structure (a validated cycle, a trajectory that has recurred
often enough to close) and \emph{collapses it into a reusable token},
quotienting a continuous cycle to a discrete point and lowering its degree,
$\kap:\Hodd\to\Heven$ \cite{li2026two}. Where $\dpar$ asks ``can I transform this toward
closure?'', $\kap$ asks ``is this already closed, and if so, can I store it as a
thing I will not have to recompute?'' This is the operator of memory\footnote{Memory is not passive storage, but the \emph{act} of converting solved computation into cheap structure.}. The Urysohn machine \cite{li2026UM} is the $\kap$-engine: it
condenses across parity sectors, moving structure from the continuous to the
discrete and stacking the resulting tokens into a ladder. Its criterion of
success is not halting but \emph{persistence}: a token is real if it recurs, if
it survives, if it earns its place by being reused.

\medskip
\noindent
The natural expectation would be that a sufficiently powerful $\dpar$-engine
suffices - that computation, given enough of it, is all intelligence needs,
with memory a mere convenience. Next, we show this expectation is false for two independent reasons. The first is a resource argument: the advice that condensation manufactures is provably not reducible to the time and space that computation spends, so no $\dpar$-engine can substitute for $\kap$ (Section~\ref{sec:resource}). The second is deeper: \emph{$\dpar$ confined to its own sector is
incomplete, and so is a pure descent dynamics confined to the other}. The
coupling is not a convenience but forced, first by resource non-fungibility and then, more fundamentally, by the incompleteness result, which generalizes G\"odel's original formulation \cite{godel1931}.

% =====================================================================
\section{Why the Coupling Is Forced: Resources and Incompleteness}
%\addcontentsline{toc}{section}{\textbf{Part II. Why the Coupling Is Forced: The Conjugate Incompleteness}}

The coupling is forced twice over, by two arguments of different depth. The
first is elementary and complexity-theoretic: the resource that condensation
manufactures is not the resource computation spends, so no amount of computation
can substitute for it (Section~\ref{sec:resource}). The second is the technical
core of this paper: each operator, confined to its own parity sector, is
provably incomplete, and the two incompletenesses are conjugate faces of a
single obstruction (Section~\ref{sec:two-incompleteness} onward). We give the
resource argument first, then the incompleteness argument.

\subsection{The resource argument: advice is a non-fungible third axis}
\label{sec:resource}

The three finitudes of Section~\ref{sec:finitudes} are naturally read as only
two resources, time and space (memory), since Turing's tradeoffs, Savitch's
included, live entirely in that plane \cite{savitch1970}. But finite time
forcing amortization and finite memory forcing abstraction both manufacture the
\emph{same} object: a structure prepared in advance so that future inference is
a lookup rather than a re-derivation. That prepared structure is a resource in
its own right, distinct from the time and space spent to build it. Borrowing the
term from complexity theory, we call it \emph{advice}: pre-positioned,
instance-independent information that a machine may consult for free at query
time \cite{arora2009computational}.

The decisive point is that advice is \emph{not fungible} with time and space.
Time and space are renewable and mutually tradeable: one can always spend more,
reuse them, or exchange one for the other, which is exactly what Savitch's
midpoint recursion does \cite{savitch1970}. Advice cannot be manufactured this way, and this is not
heuristic. Nonuniform advice is provably more powerful than unbounded ordinary
computation: the class $\mathrm{P/poly}$ contains \emph{undecidable} languages \cite{sipser_introduction_1996}.
A unary encoding of the halting problem, which no time or space budget whatsoever
can decide, is decided by a single advice bit per input length (the bit that
records whether the unique length-$n$ string lies in the language)
\cite{arora2009computational}. Advice buys what time and space, at any
price, cannot; it is a genuinely separate axis, not a repackaging of the first
two.

The new perspective yields a \emph{resource}-theoretic reading of the two operators and a first reason
the coupling is forced. Computation, $\dpar$, is the operator that \emph{spends}
time and space: it drives open structure toward closure by expending steps and
scratch memory. Condensation, $\kap$, is the operator that \emph{manufactures
advice}: it takes a completed closure and deposits it as pre-positioned
structure, the scaffold, that later inference consults for free. Amortization
(Section~\ref{sec:finitudes}) is precisely the conversion of spent time and
space into advice \cite{gershman2014amortized}, and the scaffold is the store of that advice. Because advice
is non-fungible, no $\dpar$-engine, however much time and space it is granted,
can produce what $\kap$ produces; the scaffold is not a computation run faster
but a different resource entirely. A system restricted to $\dpar$ is 
not slow but \emph{resource-incomplete}: it has no access to an axis a
finite agent in a non-stationary world cannot do without. That $\kap$ is
irreducible to $\dpar$ is reached here by an elementary accounting of resources;
we can sharpen it into the stronger statement that each
operator is not only distinct from but \emph{incomplete without} the other next.

Two caveats keep the argument honest. First, advice is not free: it must be
built, and its cost is paid in the time, space, and experience of a prior phase
(evolutionary, developmental, or a training run), which is exactly the work
$\kap$ performs over recurring experience \cite{gershman2014amortized}. The
claim is non-reducibility, not a free lunch. Second, the three-axis picture is a
conceptual decomposition, not a proved conservation law; what is rigorous is the
non-fungibility \emph{witness}, that $\mathrm{P/poly}$ decides languages no
ordinary computation can, which by itself defeats the assumption that a finite
agent has only time and space to spend. The quantitative geometry relating the
three remains open (refer to Open Problem 1 in Sec. \ref{sec:conclusion}).

\subsection{Two incompletenesses}
\label{sec:two-incompleteness}

\paragraph{The symbolic incompleteness (G\"odel)}

G\"odel's first incompleteness theorem is the canonical limit on a
$\dpar$-engine operating in the symbolic sector: any consistent, recursively
axiomatizable theory extending arithmetic contains a true sentence it cannot
prove. The mechanism is \emph{diagonalization}. Through arithmetization, the
theory is made to talk about its own provability, and the diagonal lemma
constructs a sentence that asserts its own unprovability: a fixed point of
the provability predicate that escapes the predicate's reach. The essential
ingredient is self-reference: the system must be expressive enough to encode
statements about itself. This is the incompleteness of \emph{computation}: the
$\dpar$-engine, driving conjectures toward proof within the symbolic sector,
cannot reach every truth.

\paragraph{The geometric incompleteness (Morse)}

There is a second, structurally different incompleteness, intrinsic to the
descent dynamics of a state space rather than to an external observer's power to
decide. It is a consequence of Morse theory and the limit of a pure
descent dynamics - i.e., the closure operation of the \emph{other} sector. We first
make precise the property of a state space.

\begin{definition}[Semantic complexity]
\label{def:semantic-complexity}
A closed smooth $d$-manifold $M$ is \emph{semantically complex} if it has
nonvanishing homology in some intermediate degree: $b_k(M) \neq 0$ for some $k$
with $1 \le k \le d-1$, where $b_k = \dim H_k(M)$ is the $k$-th Betti number.
Equivalently, $M$ is not a homology sphere: its topology is not exhausted by a
single connected component (degree $0$) and a single topological class (degree $d$).
\end{definition}

The term is chosen deliberately: an intermediate homology class is a feature of
the state space that is neither a discrete component (a token, degree $0$) nor
the ambient volume (degree $d$), but a genuine relational structure (e.g., a hole, a handle, a cycle) of the kind a semantically rich representation carries.
The definition is what makes the following theorem true because a
homology sphere (e.g.\ $S^d$, whose height function has exactly one minimum and
one maximum and no intermediate critical point), which is \emph{not} semantically
complex, is correctly excluded.

\begin{theorem}[Metric-topological incompleteness]
\label{thm:morse}
Let $M$ be a semantically complex closed manifold
(Definition~\ref{def:semantic-complexity}). Then:
\begin{enumerate}[leftmargin=1.9em,label=\textnormal{(\arabic*)}]
\item \textbf{Morse necessity.} Any Morse function $E:M\to\R$ possesses at least
one critical point of intermediate index $k$ with $1\le k\le d-1$. Under the
gradient flow $\dot z = -\nabla_g E$, such a point is a saddle-type
equilibrium.
\item \textbf{No-saddle obstruction.} There exists no Morse (equivalently,
structurally stable) smooth Lyapunov function on $M$ whose critical set consists
only of minima and maxima.
\item \textbf{Genericity.} For any smooth $E$, an arbitrarily small $C^2$
perturbation $\tilde E$ is Morse, and therefore must contain at least one such
saddle.
\end{enumerate}
\end{theorem}

\begin{proof}
All three clauses follow from the Morse inequalities. For a Morse function $E$
on a closed manifold, let $c_k$ be the number of index-$k$ critical points. The
weak Morse inequalities give $c_k \ge b_k$ for every $k$.

(1) By semantic complexity, $b_k \ge 1$ for some $1\le k\le d-1$; and $c_k\ge
b_k\ge 1$, so $E$ has at least one critical point of that intermediate index. An
index-$k$ critical point with $1\le k\le d-1$ has both a nonempty stable
manifold (dimension $k$) and a nonempty unstable manifold (dimension $d-k$),
i.e.\ it is a saddle of the gradient flow.

(2) A function whose critical set consists only of minima (index $0$) and maxima
(index $d$) has $c_k=0$ for all intermediate $k$. This contradicts $c_k\ge
b_k\ge 1$ from clause (1). So, no Morse function on $M$ can avoid intermediate
critical points; since a smooth Lyapunov function that is structurally stable is
Morse, no such saddle-free Lyapunov function exists.

(3) Morse functions are open and dense in $C^\infty(M)$ in the $C^2$ topology
(Morse-Sard); given any smooth $E$, an arbitrarily small $C^2$ perturbation
$\tilde E$ is Morse and so falls under clause (1).
\end{proof}

The mechanism here is \emph{not} diagonalization and not an external observer's
undecidability. It is intrinsic to the descent dynamics: the topology of the
state space forces the flow to pass through a saddle - a point that is neither
a pure minimum (it cannot simply rest) nor a pure maximum (it is not a pure
source), but a mixed equilibrium with both contracting and expanding directions.
There is no monotone descent to a clean resolution; the saddle is unavoidable.
%\paragraph{Why ``topological G\"odel theorem'' overclaims}
It is tempting to say that metric-topological incompleteness simply \emph{is} G\"odel's theorem in
geometric language. This overclaims, as we argue,
because the two obstructions have different mechanisms:

\begin{center}
\begin{tabular}{lll}
\hline
& \textbf{Symbolic(G\"odel)} & \textbf{Geometric (Morse)} \\
\hline
Engine & $\dpar$ (symbolic computation) & descent (geometric closure) \\
Self-reference & essential & absent \\
Mechanism & diagonalization & topology forces a saddle \\
Unreachable object & the G\"odel sentence & the intermediate-index saddle \\
The gap & syntax $\subsetneq$ semantics & monotone descent is impossible \\
\hline
\end{tabular}
\end{center}
\noindent
G\"odel's theorem is about self-reference creating unprovable truths within a
formal system. The Morse result is about the topology of a state space forcing
its own descent dynamics through an unresolvable saddle. These are related but
not the same, and a unification that ignores the difference explains nothing.
We propose a unification that \emph{respects} the
difference: the two are the parity-conjugate ways a single-mode resolution can
fail, which is exactly why intelligence cannot rely on a single mode.

\paragraph{The bridge: the saddle is the parity boundary}

The unifying observation is that the Morse saddle is not analogous to the
parity split of the two operators; instead, it \emph{generates} the split. At a
saddle, the gradient flow \(\dot z=-\nabla_g E\) decomposes the tangent space
into contracting and expanding directions. In the scaffold-flow language, this
is precisely the local origin of the scaffold/flow distinction. A purely
contractive system would absorb every perturbation into a stored representative;
a purely expansive system would continue to explore without committing
anything \cite{gromov2007metric}. A saddle contains both behaviors in the same local chart. Its
contracting manifold identifies the directions along which variation can be
settled, retained, and amortized as scaffold; its expanding manifold identifies
the directions along which unresolved variation remains active as flow.
Therefore, the saddle is not a topological obstruction to global descent but the local geometric event at which a finite intelligent system is forced to
separate what can be condensed from what must remain computable. The parity
boundary is produced by the dynamics whenever semantic complexity prevents all directions from being simultaneously collapsed.

\begin{thesis}[Forced saddle, forced parity boundary]
\label{thesis:saddle-parity}
On a semantically complex state space, Theorem~\ref{thm:morse} forces a saddle,
and the saddle's stable/unstable tangent splitting \emph{is} the scaffold/flow
(even/odd) decomposition realized at a point. The forced existence of the saddle
is therefore the forced existence of the parity boundary: a state space with
nontrivial intermediate homology cannot run a descent dynamics without
producing, somewhere, the contraction-expansion splitting that the parity
grading describes globally.
\end{thesis}

\subsection{The parity framework}
\label{sec:parity}

\paragraph{Coherent complexes and the parity grading}
Before assigning roles to the two operators, we first need a common substrate on
which both can act. We start with a pair of definitions to separate
\emph{coherence} from \emph{dynamics}. Coherence is supplied by a chain complex:
structures have boundaries, boundaries have no further boundary, and the law
\(\dpar^2=0\) \cite{wheeler1990information} guarantees that closed objects, holes, and conserved components
are well defined. Dynamics will enter later through the two operators, one that
moves across boundaries and the other that condenses reusable structure. The parity
grading records the minimal alternation already present in any such complex:
even degrees carry component-like, token-like structure, while odd degrees carry
cycle-like, path-like structure. %The definitions below do not yet impose the scaffold-flow interpretation; they provide the algebraic stage on which that interpretation becomes precise.

\begin{definition}[Coherent computational complex]
\label{def:complex}
A \emph{coherent computational complex} is a finite chain complex $\C =
(C_0,\dots,C_n)$ of finite-dimensional real inner-product spaces with boundary
maps $\dpar_p : C_p \to C_{p-1}$ satisfying $\dpar^2=0$, with adjoints
$\dpar^\ast$ and Hodge Laplacians $\Delta_p = \dpar_{p+1}\dpar_{p+1}^\ast +
\dpar_p^\ast\dpar_p$. The Hodge decomposition gives $C_p = \im\dpar_{p+1} \oplus
\ker\Delta_p \oplus \im\dpar_p^\ast$ with $\ker\Delta_p \cong H_p(\C)$.
\end{definition}

\begin{definition}[Parity grading]
\label{def:parity}
The degree grading induces a $\mathbb{Z}/2$-grading $\C = C_{\mathrm{even}}
\oplus C_{\mathrm{odd}}$, and $\dpar$ is the parity-reversing operator. Write
$\Heven = \bigoplus_m H_{2m}$ and $\Hodd = \bigoplus_m H_{2m+1}$.
\end{definition}

The two operators attach to the two parities at the level where the assignment
is unconditional. Degree $0$ (even) is the \textbf{symbolic register}: $H_0$
counts connected components - namely, discrete, re-identifiable tokens, the home of
$\dpar$'s formal-symbolic computation and $\kap$'s stored scaffold. Degree $1$
(odd) is the \textbf{geometric register}: $H_1$ counts cycles - namely, trajectories,
loops, paths, the home of dynamical-geometric flow.

\begin{remark}[Scope of the even/odd labeling]
\label{rem:parity-scope}
We use ``symbolic $=$ even'' and ``geometric $=$ odd'' rigorously only at the
$H_0/H_1$ boundary, where they reduce to the standard reading of components and
cycles. The claim that the alternation continues up the degrees is a modeling
hypothesis, \emph{not} a theorem of algebraic topology, and nothing below
depends on it. The load-bearing asymmetry is $H_0$ versus $H_{\ge 1}$: discrete
component-identity versus continuous cycle-traversal.
\end{remark}

\paragraph{Closure operators and their homological reach}

Each sector carries a \emph{closure operator}: the operation that drives an open
state to a resolved one. These are exactly the two operators of
Section~\ref{sec:operators}, seen as attempts at completion within a single
sector.

\begin{definition}[Sectoral closure]
\label{def:closure}
For the even (symbolic) sector, the closure operator $K_{\mathrm{sym}}$ is
\emph{proof} (the $\dpar$-engine confined to the symbolic sector): it attempts
to drive a conjecture (an open chain, $\dpar s \neq 0$) to a theorem (a closed
chain, $\dpar s = 0$) using only operations internal to the symbolic sector. For
the odd (geometric) sector, the closure operator $K_{\mathrm{geo}}$ is
\emph{gradient descent}: under the flow $\dot z=-\nabla_g E$ it attempts to drive
an arbitrary state to a clean resolution (a global minimum, a pure attractor) using only the monotone descent dynamics internal to the geometric sector.
In each case, the sector is \emph{complete} iff its closure operator resolves
every state without obstruction.
\end{definition}

\subsection{Incompleteness as non-surjective closure, and its duality}
\label{sec:duality}

The definition deliberately makes the two sectors parallel. In each case, a
closure operator is allowed to use only the resources internal to its own
sector, and completeness means that no state requiring resolution lies beyond
that internal closure. The question is
whether they fail in the same structural way. The propositions below are affirmative: symbolic closure fails because diagonalization produces a closed
class unreachable by proof, while geometric closure fails because Morse
complexity forces a critical point unreachable by pure descent to a minimum.
Incompleteness appears, in both sectors, as the failure of an internal
closure operator to cover all coherent states of the sector.

\begin{proposition}[Symbolic incompleteness as non-surjectivity]
\label{prop:godel}
$K_{\mathrm{sym}}$ is not surjective onto $\Heven$: there exist nontrivial
even-sector homology classes (true sentences, $[z]\neq 0$) not in the image of
the internal proof-closure. The obstruction is diagonalization: the diagonal
lemma constructs a closed state (the G\"odel class) that the closure operator
cannot reach from within the sector.
\end{proposition}

\begin{proposition}[Geometric incompleteness as a forced saddle]
\label{prop:geometric}
On a semantically complex state space, $K_{\mathrm{geo}}$ cannot resolve every
state by monotone descent: by Theorem~\ref{thm:morse}, any Morse energy $E$ has
an intermediate-index critical point, so the gradient flow possesses a saddle
equilibrium that is neither a minimum (a clean resolution) nor a maximum (a pure
source). The obstruction is intrinsic to the descent dynamics, and by clause~(3)
of Theorem~\ref{thm:morse} it is generic, surviving every small perturbation of
$E$.
\end{proposition}

The two propositions have the same form (a single-mode resolution is
obstructed from within its sector) but different mechanisms. The parity
framework explains why both hold, and why they are dual rather than identical.

\begin{theorem}[Parity duality of incompleteness]
\label{thm:duality}
On a coherent computational complex over a semantically complex state space, let
$K_{\mathrm{sym}}$ be symbolic proof-closure on the even sector and
$K_{\mathrm{geo}}$ be gradient descent on the odd sector. Then we have:
\begin{enumerate}[leftmargin=1.8em]
\item each closure is obstructed from within its own mode
(Propositions~\ref{prop:godel} and~\ref{prop:geometric});
\item the obstruction is parity-mixed at the saddle: an index-$k$ saddle ($1\le
k\le d-1$) has a nonempty stable subspace (even/scaffold directions) \emph{and}
a nonempty unstable subspace (odd/flow directions), so it belongs to neither
sector alone but to the boundary $\dpar$ couples;
\item consequently no single-parity (single-mode) closure can be complete:
monotone descent confined to the contracting directions stalls at the saddle
because the expanding directions remain unresolved, and symbolic closure
confined to the discrete sector stalls at the G\"odel class because reaching it
would require leaving the sector. $\dpar^2=0$, by keeping the sectors
orthogonal, is exactly what prevents a single-mode operator from resolving the
mixed obstruction.
\end{enumerate}
\end{theorem}

\begin{proof}[Proof sketch]
(i) is the two propositions. For (ii), at an index-$k$ critical point the
Hessian of $E$ has $k$ negative and $d-k$ positive eigenvalues, splitting the
tangent space orthogonally into a stable subspace $T^s$ (dimension $k$,
contracting) and an unstable subspace $T^u$ (dimension $d-k$, expanding); for an
intermediate index both are nonempty (Theorem~\ref{thm:morse}(1)). Under the
identification of contracting directions with the even (scaffold) sector and
expanding directions with the odd (flow) sector
(Thesis~\ref{thesis:saddle-parity}), the saddle carries both parities and lies
on the boundary between them. For (iii), a descent operator restricted to the
stable directions converges along $T^s$ but leaves $T^u$ unresolved, so it
cannot drive the state to a clean minimum; dually, symbolic closure restricted
to the discrete sector cannot reach the G\"odel class, whose resolution would
require the complementary parity. Orthogonality of the sectors under $\dpar^2=0$
is what makes each single-mode operator blind to the complementary subspace.
Therefore, completeness requires both parities and the boundary traffic between them,
which no single-mode operator possesses.
\end{proof}

\begin{corollary}[One fact, two faces]
\label{cor:onefact}
G\"odel incompleteness and geometric(metric-topological) incompleteness are the even and
odd faces of a single fact: \emph{no resolution confined to one mode can come
full circle.} In the even sector, this failure is realized by diagonalization
(the unprovable fixed point); in the odd sector, by the forced saddle (the
equilibrium with both contracting and expanding directions). The G\"odel
sentence is the symbolic saddle; the Morse saddle is the geometric G\"odel
sentence.
\end{corollary}

% =====================================================================
\section{The Solution: Intelligence as the Coupling}
\label{sec:solution}
%\addcontentsline{toc}{section}{\textbf{Part III. The Solution: Intelligence as the Coupling}}

\subsection{The coupling, and what it resolves}
\label{sec:coupling}

We have demonstrated that neither operator, confined to its own sector, can be complete: computation ($\dpar$)
alone hits the G\"odel class, descent alone hits the forced saddle, and these
are not two accidents but one obstruction wearing two parities. A single-mode
system (e.g., a pure symbol manipulator or a pure dynamical relaxation) is
provably insufficient for the task of intelligence \cite{bennett2023brief}. The only way to resolve a parity-mixed
obstruction is to operate in \emph{both} parities and to move structure across
the boundary between them. That cross-boundary traffic is exactly what the two
operators, coupled, provide: $\dpar$ works within a sector toward closure, and
$\kap$ carries a completed closure across the parity boundary into the
complementary sector, where it becomes a reusable token. This is the sense in
which the coupling is forced.

\begin{principle}[The coupling]
\label{prin:coupling}
Intelligence is neither $\dpar$ nor $\kap$ but their \emph{coupling}: the joint
dynamics of computation (transformation toward closure, within a sector) and
condensation (collapse of closure into a token, across sectors), on a coherent
complex under $\dpar^2=0$. The coupling is forced twice over: elementarily by the non-fungibility of advice (Section~\ref{sec:resource}), and more deeply by the incompleteness, because no single-mode closure can come full
circle, a system that must resolve parity-mixed obstructions has to run both
operators and the boundary traffic between them. Computation without
condensation is a machine that recomputes forever and never learns, and remains
trapped at the G\"odel class of its own sector; condensation without computation
is a memory that recognizes and stores but never reasons within what it has
stored. Intelligence is the traffic between them.
\end{principle}

The two operators are complementary in a way that is not symmetric. $\dpar$
works inside a token, elaborating structure toward a result; $\kap$ works
between tokens, building the next token from a completed computation. A pure
$\dpar$-agent is Turing's: powerful in principle but condemned to re-derive
everything, and against a non-stationary world it must track in bounded time, effectively helpless, because it pays the full cost of computation on every
encounter, and incomplete besides (Proposition~\ref{prop:godel}). A pure
$\kap$-agent is an inert archive: it can recognize and file but has no engine to
compute \emph{within} a filed context, so it cannot act on anything genuinely
new, and it stalls at the forced saddle (Proposition~\ref{prop:geometric}).
Only the coupling is intelligent, because only the coupling both \emph{builds}
reusable structure and \emph{computes} inside it and only the coupling spans
both parities, which Corollary~\ref{cor:onefact} shows is necessary for any
resolution to come full circle.

\subsection{What ``computation plus condensation'' means precisely}
\label{sec:comp-plus-mem}

The claim that intelligence is \emph{computation plus condensation} is easy to
state loosely and important to state sharply, because the sharp version locates
exactly what the theory of computation was missing.
The Turing machine has a tape, and one might think it therefore already has
memory. It does not, in the relevant sense. Its tape is \emph{passive}: writing
to it does not change the complexity of any future computation. A Turing machine
that has solved a problem is no faster at solving it a second time; it begins
each computation cold. This is the precise sense in which \emph{the theory of
computation cannot represent learning} \cite{bengio2013representation}. Learning \emph{is} amortization, the
conversion of solved computation into cheaper future computation, and the
Turing machine, structurally, refuses to amortize, because it has no operator
that turns a completed computation into reusable structure. %It recomputes by construction.

So the missing half is the newly defined $\kap$ operator: memorization as a \emph{first-class computational act}, the operation
of having memorized something and thereby making the next computation cheaper.
This is why the coupling is the right foundational object: the Church-Turing world formalized $\dpar$ and, because its
agent was immortal, had no reason to formalize $\kap$; it produced a
complete theory of \emph{computation} but NO theory of \emph{learning} \cite{goodfellow_deep_2016}, and the
absence was invisible precisely because the abstraction that removed it was so
successful.
We stress the register of this claim, because it is where the account could
overreach and will not. The coupling does not compute anything \emph{Turing
cannot}. A Turing machine with unbounded tape and unbounded time can compute any
computable function, condensation or no. The claim is not about
\emph{computability} but about \emph{bounded-agent efficiency}: the coupling is
what lets a finite, mortal system act well, in bounded time, on a non-ergodic
world that a pure $\dpar$-agent could only recompute its way through life expectation. Intelligence is not more powerful than
computation. It is computation made \emph{survivable} for a finite agent, and
survivability, not power, is what the extra operator buys.

\subsection{The price of the coupling: context-identification}
\label{sec:context}

The coupling of computation and condensation does not eliminate the classical
stability-plasticity dilemma \cite{grossberg1987competitive} but relocates it. Stability requires that a
recurring state be routed back to an already condensed scaffold and plasticity requires that a
genuinely novel state escape old scaffolds. The architecture needs a hinge operation: at each moment, it
must decide whether the present state is an instance of something already
known, in which case it should recollect through \(\kap\), or whether it is a
new configuration requiring active computation through \(\dpar\). This
recognize-versus-discover decision is \emph{context-identification}.
Seen this way, the stability-plasticity dilemma is not primarily a conflict
between two learning rates or two regularizers. It is a \emph{decision problem} at the boundary between stored structure and open computation \cite{green1966signal}. A false recognition
routes novelty into an old scaffold, producing rigidity, negative transfer, or
corruption of a stored token. A false discovery treats recurrence as novelty,
producing redundant tokens, fragmentation of memory, and loss of amortization.
The hardest step in the coupled architecture is neither computation nor
condensation alone, but deciding which operator should act \cite{tanner1954decision}. This is where
the symbolic and geometric incompletenesses cease to correspond and
begin to coincide: the system must decide whether the current orbit belongs to
a known basin of attraction or whether it must generate a new one.

\paragraph{Recognize-versus-discover as orbit reachability.}

Fix a state space carrying a dissipative flow
$\dot z=-\nabla_g E$,
whose attractors represent stored tokens, namely the closed structures
condensed by \(\kap\). Let \(R_k\) denote the recognition region of stored
context \(k\). These regions need not be single points; they are basins,
tubes, or neighborhoods within which the system treats a state as an instance
of an already committed scaffold.

\begin{definition}[Context-identification]
\label{def:detect}
The \emph{context-identification problem} \(\mathrm{DETECT}\) is the decision:
given a state \(z\), does its forward orbit under the flow satisfy
$\exists t\ge 0,\ \exists k
\phi_t(z)\in R_k$,
in which case the system should \emph{recognize} the state and route it to the
stored context \(k\); or does the orbit avoid all stored regions and satisfy
the novelty criterion for forming a new closed structure, in which case the
system should \emph{discover} a new context?
\end{definition}

This formulation gives the stability-plasticity dilemma a precise dynamical
meaning \cite{carpenter1987art2}. Stability is successful recognition: the orbit enters the correct
stored region and the system retrieves a scaffold already paid for by past
computation. Plasticity is successful non-recognition: the orbit fails to enter
any existing region and is allowed to close into a new token. The dilemma
arises because the two errors are dual. Expanding the regions \(R_k\) improves
recognition and protects stability, but increases the risk of absorbing novelty
into the wrong scaffold. Shrinking them protects novelty and improves
plasticity, but increases the risk of failing to recognize recurrence. The
classical tradeoff becomes a geometric boundary problem: where should the
architecture place the recognition regions, and how reliably can it decide
orbit membership near their boundaries?

\begin{proposition}[Undecidability of context-identification]
\label{prop:detect-undecidable}
If the flow is expressive enough to embed a Turing machine, which recurrent
neural dynamics are~\cite{siegelmann1995}, then $\mathrm{DETECT}$ is
undecidable: no algorithm decides, for arbitrary $z$, whether $\phi_t(z)$
eventually enters a stored region.
\end{proposition}

\begin{proof}
$\mathrm{DETECT}$ is an instance of orbit reachability: does the orbit of $z$
enter a designated region? For dynamical systems capable of simulating a Turing
machine, reachability of a region is undecidable, being interconvertible with
the halting problem~\cite{moore1990}; the halting configuration is encoded as
the target region, and the orbit reaches it iff the simulated machine halts.
Embedding such a machine in the flow (available by~\cite{siegelmann1995})
transports the undecidability to $\mathrm{DETECT}$. Eventual entry into $R_k$ is
a statement about unbounded time, and no bounded-time procedure decides it in
general.
\end{proof}

This places $\mathrm{DETECT}$ on the geometric side: it is orbit-reachability,
undecidable for the same reason the halting problem is. But orbit-reachability
alone would make it undecidable in the generic way many dynamical questions are.
What makes it \emph{diagonally} undecidable, in the self-referential manner of
G\"odel and Turing, is a second feature.

\paragraph{The diagonal: DETECT decides membership in a structure it builds}

The stored regions $\{R_k\}$ are not given in advance. They are the tokens that
past $\mathrm{DETECT}$ decisions created: each ``discover'' outcome condenses a
new cycle into the scaffold and thereby defines a new $R_k$. Therefore,
$\mathrm{DETECT}$ closes a diagonal loop by deciding membership in a structure that it is simultaneously constructing. 

\begin{proposition}[Diagonal obstruction]
\label{prop:diagonal}
No total decider for $\mathrm{DETECT}$ is correct on all inputs whose
context-membership is permitted to depend on the decider's own output.
\end{proposition}

\begin{proof}
Suppose $D$ decides $\mathrm{DETECT}$. Construct an input $z^\ast$ whose
generating process is: run $D(z^\ast)$; if $D$ answers ``recognize'' (old
context), steer the flow into a fresh region disjoint from every stored $R_k$
(making the truth ``discover''); if $D$ answers ``discover'' (new context),
steer the flow into an existing $R_k$ (making the truth ``recognize''). The
steering is realizable because the flow is Turing-capable and can condition its
trajectory on the simulated value $D(z^\ast)$. Then $D$ is wrong on $z^\ast$ by
construction. This is the halting-problem diagonalization with
``recognize/discover'' in place of ``halt/loop.''
\end{proof}

Propositions~\ref{prop:detect-undecidable} and~\ref{prop:diagonal} exhibit both
incompletenesses at a single operation. The recognize/discover
boundary is a \emph{separatrix} of the flow - the stable manifold of a saddle
(Theorem~\ref{thm:morse}), dividing the basin of a stored token from the basin
of a new one, so deciding which side of it a state lies on is the
forced-saddle (geometric) incompleteness; a state exactly on the separatrix is
undecidable in the limit. Simultaneously, $\mathrm{DETECT}$ deciding membership
in the scaffold it builds is the diagonal (symbolic) incompleteness.

\begin{corollary}[Coincidence of the incompletenesses]
\label{cor:coincidence}
Context-identification is undecidable for both parity-conjugate reasons of
Corollary~\ref{cor:onefact} at once: geometrically, it decides the side of a
saddle separatrix; symbolically, it decides membership in a self-referentially
constructed structure. It is the operation at which the diagonal and the forced
saddle coincide rather than correspond at the coupling's hinge (its
hardest point).
\end{corollary}

\paragraph{The mechanism: scale-recursion, and the irreducible floor}

Why should a physical system's context-identification actually encounter the
undecidable instances? Because whether two fine-scale tokens are ``the same
context'' is scale-dependent \cite{anderson1972more}: distinct at level $\ell$, they may merge into one
super-context at level $\ell{+}1$. Deciding context-identity at a fine scale can
require the coarse-grained landscape one level up, which exists only after
level $\ell$ has been condensed. Read as a renormalization-group flow \cite{wilson1974renormalization}, the
``true'' context is a fixed point of coarse-graining, and the required depth to
reach it is not bounded a priori. Ladder depth plays the role that unbounded
time plays in the halting problem \cite{li2026UL}.

\begin{proposition}[Depth regress]
\label{prop:depth}
Let $\mathrm{DETECT}_d$ be context-identification restricted to a scale-depth
budget $d$. Then $\mathrm{DETECT}_d$ is decidable for each fixed $d$, at cost
growing with $d$; but the depth $d^\ast(z)$ required for the answer to stabilize
is not bounded a priori, and deciding $d^\ast(z)$ in advance is equivalent to
$\mathrm{DETECT}$ itself.
\end{proposition}

The consequence is a prediction instead of a caveat. Any bounded-time,
bounded-depth learner computes a bounded-depth approximation to an undecidable
predicate, and therefore carries an \emph{irreducible} indexing error rate that
no refinement removes. That error is not uniform: it concentrates where the
undecidability bites, near separatrices (ambiguous boundary states) and for
inputs whose context-identity stabilizes only at large depth. Intelligence is
the never-finished response of a finite agent to an inexhaustible world; the
irreducibility of its central error is the mark of the inexhaustibility. The
coupling does not \emph{remove} the incompleteness that forced it into being because it is the optimal management of an obstruction that cannot be removed.

\section{Implications: Factorized Intelligence via Structural Decoupling}

\subsection{Structural decoupling: the scaffold-flow factorization}
\label{sec:decoupling}
 
The coupling of the two operators is not a pairing of two processes that
happen to run side by side. The coherence law $\dpar^2=0$ that forces both
operators to exist also forces their targets to be \emph{orthogonal}, and this
orthogonality, which we call \emph{structural decoupling}, is what makes
the coupling implementable on a single finite substrate. The Hodge
decomposition of Definition~\ref{def:complex} splits every degree orthogonally,
$C_p \;=\; \underbrace{\im\dpar_{p+1}\oplus\im\dpar_p^\ast}_{\text{flow}}
\;\oplus\; \underbrace{\ker\Delta_p}_{\text{scaffold}}$,
into an exact/co-exact part carrying the \emph{metric variation} within a
context (the \emph{flow}) and a harmonic part carrying the \emph{topological
invariants} that identify the context (the \emph{scaffold}, $\ker\Delta_p\cong
H_p$). The two are orthogonal by construction: the scaffold is precisely what
boundary and co-boundary cannot reach, the flow is precisely their image.
Orthogonality is not an added assumption but the Hodge theorem applied to a
complex satisfying $\dpar^2=0$.
 
\begin{principle}[Structural decoupling]
\label{prin:decoupling}
On a coherent complex, the coherence law $\dpar^2=0$ forces an orthogonal
factorization of every representation into a \emph{topological scaffold} (the
harmonic sector: discrete, invariant, which-context) and a \emph{metric flow}
(the exact/co-exact sector: continuous, variable, within-context). The two are
decoupled: a change confined to the flow does not alter the scaffold, and vice
versa. Intelligence is carried on this factorization, $\kap$ builds and
maintains the scaffold, $\dpar$ computes within the flow, and the
orthogonality is what lets a single finite substrate hold both without
interference.
\end{principle}
 
We name the factorization itself, since it is the object a learning theory built
on this foundation actually manipulates.
 
\begin{definition}[Metric-topological factorization]
\label{def:mtf}
The \emph{metric-topological factorization} (MTF) of a representation is its
orthogonal decomposition into scaffold and flow: a discrete topological
component (invariant, reusable, identifying \emph{which} context) and a
continuous metric component (variable, refined, predicting \emph{within} the
context). Adaptive inference factorizes accordingly, into which-context
inference (a scaffold read) and within-context prediction (a flow computation).
\end{definition}
 
Structural decoupling redeems the promises of Section \ref{sec:problem}. The stability-plasticity
problem, how a finite substrate can be at once adaptive and retentive, is
resolved because scaffold and flow are orthogonal: the flow can be updated
(plasticity) without disturbing the scaffold (stability), so the sector-partition
that finitude forced (Section~\ref{sec:finitudes}) is realized concretely as the
Hodge splitting \cite{lim2020hodge}. Generalization and memorization become
orthogonal likewise: generalization is extraction of the scaffold (the invariant
that recurs across episodes), memorization is retention of the flow (the
idiosyncratic specifics of one episode). Because the two are decoupled, an agent
can keep the scaffold and discard the flow, generalize without memorizing,
and overfitting is exactly the \emph{collapse} of the factorization, flow content
wrongly stored as though it were scaffold. The parity split, the resolution of
stability-plasticity, and generalization-without-memorization are three faces of
one fact: the coherence law decouples the metric from the topological.
 
\subsection{Structural learning theory}
\label{sec:slt}
 
If a representation factors into scaffold and flow, then \emph{learning} is not
parameter fitting but the construction and maintenance of that factorization.
This is the organizing claim of what we call \emph{structural learning theory}:
the study of learning as structural decoupling under the coherence law.
To learn, on this account, is to do two things, one per operator. To \emph{build
the scaffold} is to discover the topological invariants of a domain, the
which-context structure, the reusable tokens, by condensing recurring flow
cycles into harmonic generators; this is $\kap$. To \emph{compute the flow} is to
calibrate the metric prediction within a fixed scaffold cell, the
within-context refinement, by transformation toward closure; this is $\dpar$.
A learner is complete when it has both discovered the scaffold that organizes its
experience and can compute the flow within each cell; and it is the
\emph{decoupling} that makes the two learnable separately, so that discovering a
new context does not overwrite the predictions inside old ones, and refining a
prediction does not move the context boundaries.
 
The recursion of Section~\ref{sec:finitudes} reappears here as a \emph{tower of
factorizations}. A scaffold at one level is flow relative to a coarser scaffold
above it: the tokens that are invariant which-context structure at level $\ell$
are the variable within-context material that a level-$(\ell{+}1)$ scaffold
organizes. Learning climbs this tower by iterated $\kap$, each rung factoring the
rung below into a coarser scaffold and a residual flow. This is the same ladder
finitude forced, complexity serialized into depth, now read as a recursion
of metric-topological factorizations, and it is the structural content of
learning a domain deeply: to have factored it, and re-factored its factors, until
what recurs has been separated from what varies at every scale.
Structural learning theory is the study of intelligence as the coupled
$(\dpar,\kap)$ dynamics \emph{viewed through the factorization it maintains}. Its
objects are scaffolds and flows rather than parameters and losses; its notion of
generalization is scaffold-survival rather than low training error; its notion of
catastrophic forgetting is factorization-collapse rather than weight overwriting;
and its central difficulty is the one Section \ref{sec:solution} identified - that deciding which
scaffold cell a state belongs to is context-identification, undecidable, so the
factorization can never be maintained without an irreducible rate of
misassignment.

% =====================================================================
\section{Conclusion}
\label{sec:conclusion}

The theory of computation is the theory of what can be calculated in the absence
of resource constraints. It was built, deliberately and successfully, around an
agent with infinite time, unbounded memory, no body, and no stake in the
outcome. Intelligence is the theory of computation \emph{under} the constraints
that theory removed, and those constraints force a second operator into
existence: alongside computation ($\dpar$), which transforms structure toward
closure, memorization ($\kap$), which condenses closure into reusable tokens.
The central result of this paper is that the coupling of the two is not a
convenience but a necessity, and necessary for a provable reason. Neither
operator alone can come full circle: computation confined to the symbolic sector
meets G\"odel's diagonalization, descent confined to the geometric sector meets
the Morse forced saddle, and these are the parity-conjugate faces of a single
obstruction that $\dpar^2=0$ imposes. A single-mode system is therefore provably
incomplete, and intelligence must couple both modes and the boundary traffic
between them. The price is exact: the coupling's hinge, context-identification,
is where both incompletenesses coincide, undecidable and carrying an
irreducible error floor: the coupling manages the obstruction that forced it,
but cannot dissolve it.
What is universal is neither an agent nor a solution but the coupled operation
itself, realized wherever a finite system meets an open world. Because it is one
law meeting ever more matter, it appears at every scale of aggregation in an
emergently distinct form, selection in life, plasticity in nervous systems,
learning in cognition, tradition in culture, one operation seen through a
stack of broken symmetries, climbing irreversibly from genes to thoughts to the
shared structures of civilizations. Intelligence, on this account, is a natural
kind in the way crystallization and combustion are natural kinds: a single
process that occurs wherever its precondition is met, at whatever scale, in
whatever substrate. Its precondition is finitude against inexhaustibility, and
its signature, the mark that it faces a world it can never close, is that
its central act, deciding what it is looking at, can never be made error-free.
Intelligence is the never-finished condensation, by a finite thing, of an
endless world.

\begin{researchproblem}[Geometry of the three resources]
\label{rp:three-resources}
Give operational measures of time, space, and advice for a coupled
$\dpar$-$\kap$ system, and characterize the achievable region among them. The
non-fungibility of advice ($\mathrm{P/poly}\not\subseteq\mathrm{Decidable}$)
fixes one boundary; a quantitative law relating advice built by $\kap$ to the
time and space saved by $\dpar$ at query time is not known.
\end{researchproblem}

\bibliographystyle{plain}
\bibliography{ref,references}

\end{document}